\documentclass[10pt,twocolumn,letterpaper]{article}

\usepackage{wacv}
\usepackage{times}
\usepackage{epsfig}
\usepackage{graphicx}
\usepackage{amsmath}
\usepackage{amssymb}

\usepackage{caption}
\usepackage{subcaption}

\usepackage{xifthen}
\newcommand{\netName}[1][]{%
    \ifthenelse{\equal{#1}{}}{DRNet}{DRNets}%
}

\usepackage{xifthen}
\newcommand{\supernet}[1][]{%
    \ifthenelse{\equal{#1}{}}{RouterNet}{RouterNets}%
}

\usepackage{float}
\floatstyle{plaintop}
\restylefloat{table}

\usepackage{array}
\newcolumntype{L}[1]{>{\raggedright\let\newline\\\arraybackslash\hspace{0pt}}m{#1}}
\newcolumntype{C}[1]{>{\centering\let\newline\\\arraybackslash\hspace{0pt}}m{#1}}
\newcolumntype{R}[1]{>{\raggedleft\let\newline\\\arraybackslash\hspace{0pt}}m{#1}}

\usepackage{tabu}
\usepackage{enumitem}
\usepackage{multicol}

\usepackage{multirow}

\newcommand{\thickhline}{%
    \noalign {\ifnum 0=`}\fi \hrule height 1pt
    \futurelet \reserved@a \@xhline
}

\newcommand{\powerpoint}[1]{\noindent\textbf{#1}}

\usepackage[export]{adjustbox}

\usepackage{dsfont}

\usepackage{flushend}

\def\wacvPaperID{633} 

\wacvfinalcopy 
\ifwacvfinal
\def\assignedStartPage{9876} 
\fi

\ifwacvfinal
\usepackage[breaklinks=true,bookmarks=false]{hyperref}
\else
\usepackage[pagebackref=true,breaklinks=true,colorlinks,bookmarks=false]{hyperref}
\fi

\ifwacvfinal
\else
\fi

\begin{document}

\title{Dynamic Routing Networks}


\author{Shaofeng Cai, Yao Shu, Wei Wang, Beng Chin Ooi\\
National University of Singapore\\
{\tt\small \{shaofeng, shuyao, wangwei, ooibc\}@comp.nus.edu.sg}
}


\maketitle

\begin{abstract}

The deployment of deep neural networks in real-world applications is mostly restricted by their high inference costs.
Extensive efforts have been made to improve the accuracy with expert-designed or algorithm-searched architectures.
However, the incremental improvement is typically achieved with increasingly more expensive models that only a small portion of input instances really need.
Inference with a static architecture that processes all input instances via the same transformation would thus incur unnecessary computational costs.
Therefore, customizing the model capacity in an instance-aware manner is much needed for higher inference efficiency.
In this paper, we propose \textit{Dynamic Routing Networks} (\netName[s]), which support efficient instance-aware inference by routing the input instance to only necessary transformation branches selected from a candidate set of branches for each connection between transformation nodes.
The branch selection is dynamically determined via the corresponding branch importance weights, which are first generated from lightweight hypernetworks (\textit{\supernet[s]}) and then recalibrated with Gumbel-Softmax before the selection.
Extensive experiments show that \netName[s] can reduce a substantial amount of parameter size and FLOPs during inference with prediction performance comparable to state-of-the-art architectures.



\end{abstract}
\section{Introduction}

Deep convolutional neural networks (CNNs)~\cite{he2016deep,zoph2018learning} have revolutionized computer vision with increasingly large and sophisticated architectures.
The architectures are typically designed and tuned by domain experts with rich engineering experience.
These models achieve remarkable performance with hundreds of layers and millions of parameters, which however consume a substantial amount of computational resources during inference.
Improving inference efficiency has thus become a major issue for the deployment of deep neural networks in real-world applications.

Recently, there has been a growing body of research on efficient network design~\cite{howard2017mobilenets,sandler2018mobilenetv2,wang2018skipnet} for more efficient inference architectures.
These works mainly focus on designing more efficient transformations to reduce the parameter size and inference FLOPs.
Many architectures with efficient transformation building blocks have been proposed.
Particularly, SqueezeNet~\cite{iandola2016squeezenet} reduces the parameter size significantly with the squeeze-and-expand convolution block.
MobileNets~\cite{howard2017mobilenets} substantially reduce the parameter size and computation cost measured in FLOPs on mobile devices with depthwise separable convolution.
Subsequent works such as MobileNetV2~\cite{sandler2018mobilenetv2} and ShuffleNetV2~\cite{ma2018shufflenet} further reduce FLOPs on the target hardware.
However, it is well recognized that designing these architectures is a non-trivial task that requires engineering expertise.

\begin{figure*}[t!]
    \centering
    \begin{subfigure}[t]{0.4\textwidth}
        \centering
        \includegraphics[height=0.45\textwidth]{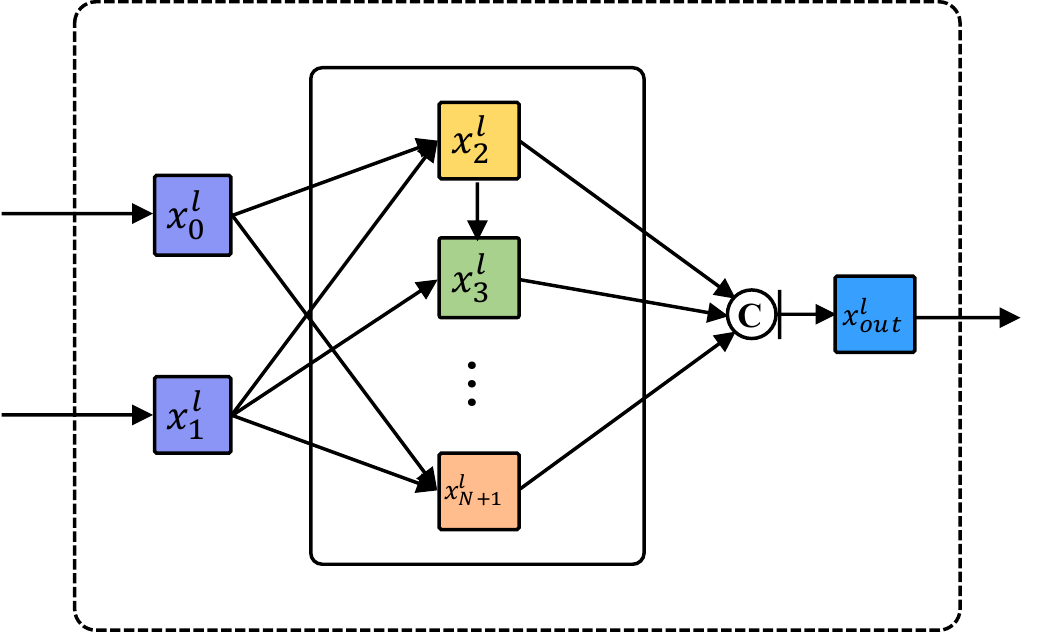}
        \caption{Cell Structure}
        \label{fig:cell}
    \end{subfigure}%
    ~ 
    \begin{subfigure}[t]{0.4\textwidth}
        \centering
        \includegraphics[height=0.5\textwidth]{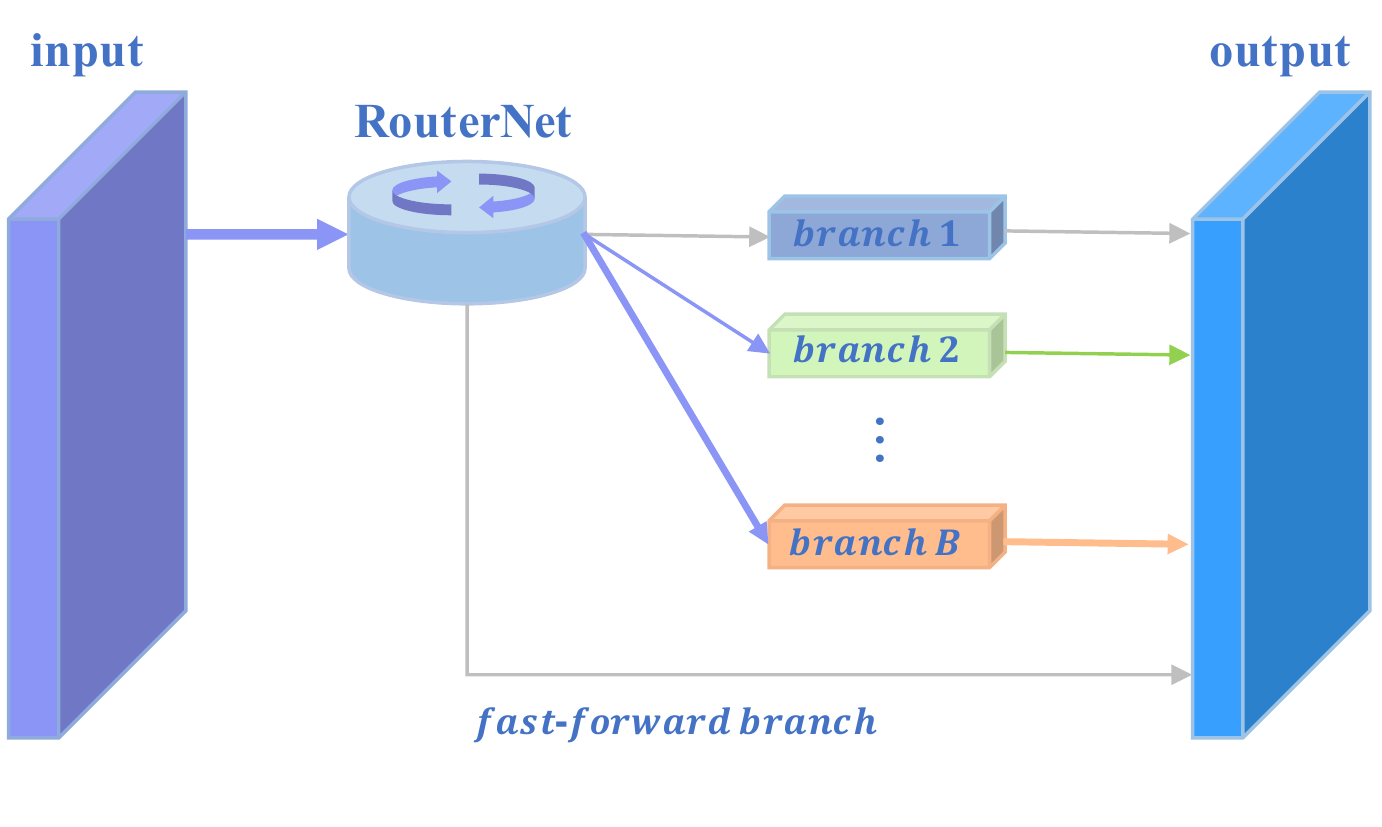}
        \caption{Node Connection}
        \label{fig:inference}
    \end{subfigure}
    \caption{\textit{Left}: the \textit{Cell} structure of \netName[s] with inter-connected nodes.
    \textit{Right}: the illustration of a node connection with $B$+1 transformation branches.
    During inference, each input instance is routed to only necessary branches for efficiency.
    Importance weights of branches are dynamically generated from the \supernet[] and then recalibrated with Gumbel-Softmax.}
    \label{fig:overall_structure}
\end{figure*}

To automate the architecture design process, there has been an increasing interest in Neural Architecture Search (NAS).
Mainstream NAS algorithms~\cite{zoph2016neural,zoph2018learning,real2019regularized} search for the network architecture iteratively. In each iteration, an architecture is proposed by the search algorithm and then trained and evaluated. The evaluation performance is in turn utilized to update the algorithm.
This process is incredibly slow because both the algorithm and the proposed architectures require extensive training.
In particular, the reinforcement learning (RL) based algorithm NASNet~\cite{zoph2018learning} takes 1800 GPU days, and the evolution based algorithm AmoebaNet~\cite{real2019regularized} costs 3150 GPU days to obtain the final architecture.
To expedite the search process, many acceleration methods~\cite{baker2017accelerating,liu2018progressive,bender2018understanding,pham2018efficient} have been proposed.
Many recent works~\cite{liu2018darts,xie2018snas,cai2018proxylessnas,wu19fbnet} instead remove the search algorithm and instead optimize parameters of both the architecture and the selection process concurrently with gradient-based algorithms.
Among these methods, the gradient-based NAS algorithms turn out to be both efficient and effective for the architecture search.

By and large, expert-designed and NAS-searched models are quite efficient and accurate during inference.
Nonetheless, most architectures of these models are static during inference, and thus not adaptive to the varying complexity of input instances.
In real-world applications, only a small portion of input instances require deep representations~\cite{wang2018skipnet,huang2017multi}, as made evident by the diminishing marginal returns of increasing model size on accuracy.
Therefore, extensive computational resources could be wasted if all input instances are processed in the same way.
To further improve the inference efficiency over existing architectures, designing a dynamic architecture with adequate representational power to support the inference of hard instances, and meanwhile a flexible mechanism to provide only necessary computation for instances of varying difficulty is thus much needed.

To support dynamic inference, we propose \textit{\netName[s]} to support instance-aware inference with the building block \textit{Cell} and the transformation of the \textit{node connection} illustrated in Figure~\ref{fig:overall_structure}.
At a high level, \textit{\netName[s]} can be regarded as a dynamic architecture generator with its backbone network building upon best-performing expert-designed and NAS-searched architectures, which produce architectures customized for the current input instance for higher inference efficiency.
Specifically, the backbone network is a stack of $L$ structurally identical \textit{Cell}s.
Each cell contains $N$ inter-connected \textit{Node}s and receives inputs from the outputs of its two preceding cells.
Instead of painstakingly searching for the connection topology and the respective transformation for each node connection of the cell as in NAS~\cite{zoph2016neural,zoph2018learning,real2019regularized,liu2018darts}, each node of \netName[s] is simply connected to a predetermined subset of its previous node(s), e.g., two in Figure~\ref{fig:cell}, and each connection transforms the current instance via a \textit{candidate operation set} of $B$+1 transformation branches.
These $B$ branches can be homogeneous or heterogeneous, which are trained to specialize in certain types of instances.
There is another fast-forward branch integrated to efficiently forward the current instance without heavy computation.

To enable instance-aware architecture customization, we integrate $L$ lightweight hypernetworks \textit{\supernet[s]}, one for each cell to dynamically determine the relative importance weight of each branch among the $B$+1 branches of each connection.
We further introduce Gumbel-Softmax~\cite{jang2016categorical,maddison2016concrete} to recalibrate the importance weights of each connection so that during training, these weights are dense and the entire architecture is efficiently optimized, and during inference, the input instance is routed only to necessary branches of larger importance weights for efficiency as illustrated in Figure~\ref{fig:inference}.
The advantages of \netName[s] can be summarized as follows:
\begin{itemize}

    \item The architecture of \netName[s] is general and customizable, which supports the instance-aware dynamic routing mechanism and thus improves inference efficiency significantly by reducing redundant computation.
    
    \item \netName[s] introduce Gumbel-Softmax to lightweight \supernet[s] during the branch selection process, which enables direct gradient descent optimization and is more tractable than the RL-based methods.
    
    \item \netName[s] achieve state-of-the-art inference efficiency in terms of parameter size and FLOPs and inherently supports applications that require runtime control.
    
\end{itemize}

Our experiments show that \netName[s] are extremely efficient during inference which selects only necessary branches on a per-input basis.
For instance, \netName[s] reduce the parameter size by 3.71x and FLOPs by 4.53x respectively compared with the NAS-searched high-performance architecture DARTS~\cite{liu2018darts} on CIFAR-10 with comparable accuracy.
With a tiny model of 0.57M parameters, \netName[s] achieve much better accuracy while with only 13.48\% parameter size and 46.41\% FLOPs during inference compared with the expert-designed network MobileNetV2 1.5x~\cite{sandler2018mobilenetv2}.
We also provide ablation studies and visualize the branch selection process to better understand the proposed dynamic routing mechanism.

The remainder of this paper is organized as follows:
we first discuss the related works in Section~\ref{sec:relatedwork}.
Then in section~\ref{sec:method}, we introduce the backbone network and the hypernetworks of \netName[s] in detail.
Experimental evaluations of \netName[s] are provided in Section~\ref{sec:experiment}, where the main results and findings are summarized in Section~\ref{subsec:evaluation} and Section~\ref{subsec:visualization}.
Finally, Section~\ref{sec:conclusion} concludes the paper.

\section{Related Work}
\label{sec:relatedwork}

\noindent
\textbf{Efficient Architecture Design.}
An increasing number of works focus on directly designing resource-aware networks~\cite{iandola2016squeezenet,ma2018shufflenet,howard2017mobilenets,sandler2018mobilenetv2}.
SqueezeNet~\cite{iandola2016squeezenet} proposes the fire module of the squeeze-and-expand convolution, which reduces the parameter size substantially.
MobileNets~\cite{howard2017mobilenets,sandler2018mobilenetv2,andrew19searching} series architectures adopt depthwise separable convolution for efficient inference.
MobileNetV2~\cite{sandler2018mobilenetv2} achieves higher efficiency with inverted residual blocks, and MobileNetV3~\cite{andrew19searching} further improves efficiency with the integration of components searched by NAS.
ShuffleNet~\cite{zhang2018shufflenet} introduces the lightweight channel shuffle operation for information exchange between channel groups, and ShuffleNetV2~\cite{ma2018shufflenet} further improve inference speed by considering the actual overhead on the target platform.
To provide efficient inference, many of these transformations are adopted in the candidate operation set of \netName[s].

\noindent
\textbf{Neural Architecture Search.}
There has been an increasing interest in 
automated architecture search (NAS).
Given the learning task, NAS aims to find the optimal architecture, specifically, the topology of network connections between nodes and the transformation operation for each connection.
NAS typically consists of two stages, i.e., architecture optimization and parameter optimization.
Mainstream NAS methods~\cite{zoph2018learning,real2019regularized} consider architecture optimization as a stand-alone process that is separated from the parameter optimization.
Search algorithms, such as RL-based NAS~\cite{zoph2016neural,zoph2018learning} and evolutionary-based NAS~\cite{xie17genetic,real2019regularized}, obtain high-performing architectures while at the cost of an unprecedented amount of search time.
Many works have been proposed to expedite the search process, e.g., via performance prediction~\cite{baker2017accelerating,liu2018progressive}, hyperparameter generating initialization weights~\cite{brock18smash}, weight sharing~\cite{pham2018efficient}, and etc.

Many recent works~\cite{cai2018proxylessnas,wu19fbnet} instead integrate the architecture search and architecture optimization into one gradient-based optimization framework.
In particular, DARTS~\cite{liu2018darts} relaxes the discrete search space to continuous operation mixing weights for each connection and optimizes the weights directly with gradient back-propagated from the validation loss.
Likewise, SNAS~\cite{xie2018snas} models the discrete search space with sets of one-hot random variables, which are made differentiable by sampling from continuous \textit{Concrete Distribution}~\cite{jang2016categorical,maddison2016concrete}.
\netName[s] also relax the discrete branch selection to continuous importance weights that are optimized by gradient descent. While instead of obtaining a set of fixed weights, \supernet[s] are introduced to dynamically generate these weights to support instance-aware architecture customization for efficiency.


\noindent
\textbf{Dynamic Inference Architecture.}
A growing body of research has been investigating methods to accelerate inference, e.g., model compression via weight pruning~\cite{song15learning}, vector quantization, knowledge distillation~\cite{hinton15distilling} and etc.
Typically, these methods are designed as post-training techniques.
Relatively fewer works explore dynamic inference architectures, which supports instance-aware execution.
The dynamic inference is based on the observation that most input instances can be accurately processed with a small network, and only a few inputs need an expensive model.

The idea of dynamic inference is related to mixture-of-experts~\cite{olnn}, whereas in \netName[s], the input is dynamically routed to specialized branches instead of individual models.
Most prior dynamic inference architectures~\cite{olnn,routingnet,modularNet} focus mainly on accuracy, and their architectures are introduced in abstract terms.
To improve efficiency, the two ResNet~\cite{he2016deep} based architectures SkipNet~\cite{wang2018skipnet} and AIG~\cite{aig} propose to dynamically skip each residual layer.
While the backbone network of \netName[s] adopts the topology and efficient branches of best-performing architectures~\cite{sandler2018mobilenetv2,liu2018darts}, which provides more efficient and diversified branch selection for higher efficiency.
Further, the dynamic skipping of SkipNet~\cite{wang2018skipnet} and AIG~\cite{aig} are based on a hypernetwork trained with policy gradient~\cite{policygradient} and straight-through Gumbel-Softmax~\cite{jang2016categorical} respectively.
However, training with policy gradient is brittle and often stuck in poor optima; the Straight-Through gradient estimation of AIG is biased~\cite{aig,jang2016categorical,maddison2016concrete} and 
easily falls into the selection collapse of sampling only a few branches repeatedly. 
Compared with SkipNet and AIG, the hypernetworks of \netName[s] are optimized more effectively 
with a two-stage temperature annealing scheme, 
and \netName[s] support a varied branch selection of multiple branches for each connection instead of the binary skipping choice.
MSDNet~\cite{huang2017multi} inserts multiple classifiers into a multi-scale version of DenseNet~\cite{huang2017densely} and supports faster predictions by exiting into a classifier.
\netName[s] can also selectively route the input to lightweight branches for accelerating predictions.

\section{Dynamic Routing Network}
\label{sec:method}

We aim to devise an efficient and flexible backbone network for \netName[] so that during inference, accompanying hypernetworks can dynamically produce a customized architecture for each input instance for higher inference efficiency.
We introduce the backbone network in Section~\ref{sec:sec_backbone}, hypernetworks in Section~\ref{sec:sec_supernet}, and the optimization and efficient inference in Section~\ref{sec:sec_optimization}.

\subsection{The Backbone Network}
\label{sec:sec_backbone}

Following NAS~\cite{zoph2016neural,liu2018darts,xie2018snas}, the backbone of \netName[] is constructed with a stack of $L$ \textit{Cells}.
As illustrated in Figure~\ref{fig:cell}, each cell is a directed acyclic graph consisting of an ordered sequence of $N$ intermediate nodes, which receives inputs $x_0^l$ and $x_1^l$ from outputs of the two preceding cells.
Each intermediate node $x_i^l (i > 1)$ of the $l_{th}$ cell learns latent representations and receives inputs from a set of $n$ previous nodes of smaller indexes $I_i^l$, e.g., $n=2$\footnote{$n$=1 and $I_i^l = \{i-1\}$ forms canonical feed-forward networks, and $n$ can be larger than two for larger model capacity and better representation learning.} in Figure~\ref{fig:cell}:

\begin{equation}
    \begin{aligned}
        x_i^l = \sum_{j \in I_i^l} \mathcal{F}_{j,i}(x_j^l), j < i \wedge |I_i^l|=n
    \end{aligned}
    \label{eq:node_op}
\end{equation}

\noindent
The main task of NAS is to find the best cell connection topology given by $I_i^l$, i.e., best previous nodes (two nodes for most NAS architectures~\cite{zoph2016neural,liu2018darts,xie2018snas}) for each intermediate node, and meanwhile, determine the best branch from candidates for each node connection.
We sidestep the architecture search by using a hypernetwork, which will be introduced in Section~\ref{sec:sec_supernet}, to dynamically determine the necessary branches for each connection given the current input instance for higher efficiency.

Thereby, there are $C=n \cdot N$ connections for each cell.
As is illustrated in Figure~\ref{fig:inference}, the connection passes information from node $x_j^l$ to $x_i^l$ via a predetermined set of $B$+1 candidate transformation branches, e.g., operations adopted from efficient networks~\cite{sandler2018mobilenetv2,zhang2018shufflenet} and NAS~\cite{liu2018darts,xie2018snas}:

\begin{equation}
    \begin{aligned}
        \mathcal{F}_{j,i}(x_j^l) = \sum_{b=0}^{B}w_b \cdot \mathcal{F}_b(x_j^l) = \sum_{b=0}^{B} w_b \cdot (\mathcal{W}_b \ast x_j^l)
    \end{aligned}
    \label{eq:connection_op}
\end{equation}

\noindent
where $\mathcal{W}_b$ is the transformation weights (e.g., the weight matrix of a linear layer or a convolutional kernel), $\ast$ denotes the transformation (e.g., matrix multiplication or a convolutional operation), and $w_b$ indicates and importance of the $b_{th}$ branch.
The branch importance weight $w_b$ is dynamically determined by the cell hypernetwork (\textit{\supernet}$^l$), rather than a fixed learned parameter as in most NAS methods~\cite{liu2018darts,xie2018snas}.
For each connection, the $B$+1 importance weights are recalibrated with Gumbel-Softmax so that these branches can be effectively trained and specialized for different input instances during training; while during inference, only the most important branches are selected for prediction efficiency.
We will elaborate on \supernet[] and Gumbel-Softmax in Section~\ref{sec:sec_supernet}.

The candidate operation set should contain at least one fast-forward branch, e.g., identity mapping, to minimize unnecessary computation for easy instances.
Further, the training cost of Equation~\ref{eq:connection_op} can be greatly reduced by preparing the aggregate transformation weights before the computation-heavy input transformation, with the reformulation: $\sum_{b=0}^{B} w_b \cdot (\mathcal{W}_b \ast x_j^l) = (\sum_{b=0}^{B} w_b \cdot \mathcal{W}_b) \ast x_j^l$.
Finally, the output of the cell $x_{out}^{l}$ is collected by \textit{concatenating} outputs from all \textit{intermediate} nodes $\{x_2^l, \dots, x_{N+1}^l\}$.
We shall use superscript $l$, subscript $c$ and $b$ to index the cell, connection, and branch respectively.

Recent works~\cite{xie2019exploring} demonstrate that architectures with randomly generated connections achieve surprisingly competitive results compared with the best-performing NAS models, which is also confirmed in our experiments.
In this paper, we thus adopt a simple connection topology introduced in Section~\ref{sec:archi_details} and focus mainly on the branch selection mechanism and the impact of dynamic routing on inference efficiency.
With such a dynamic architecture, we can readily adjust the candidate operation set
for each connection to customize the model capacity and inference efficiency based on the task difficulty and resource constraints in deployment.

\subsection{Dynamic Weight Recalibration with RouterNet}
\label{sec:sec_supernet}

To support instance-aware inference control, we introduce one lightweight hypernetwork \textit{\supernet} for each cell.
Each $\textit{\supernet[]}^l$ receives the same input as the $l_{th}$ cell, specifically, the output nodes $x_{out}^{l-2}, x_{out}^{l-1}$ ($x_{0}^{l}, x_{1}^{l}$) from the two preceding cells, and generates importance weights $\boldsymbol{W}^l$ for the $C$ connections of the corresponding cell at once:

\begin{equation}
    \begin{aligned}
        \boldsymbol{W}^l = \textit{\supernet}^l (x_{0}^{l}, x_{1}^{l}), \boldsymbol{W}^l \in \mathbb{R}^{C \times (B+1)}
    \end{aligned}
    \label{eq:supernet_op}
\end{equation}

\noindent
Specifically, each $\textit{\supernet[]}^l$ weighs each branch of the connection with the respective importance weight during training; and further, as illustrated in Figure~\ref{fig:inference}, routes the current input to only necessary branches during inference.
In this work, the \supernet[] comprises a pipeline of 2 convolutional blocks, a global average pooling, and finally an affine transformation to produce the weights.
The convolution block adopts the separable convolution~\cite{sandler2018mobilenetv2}, specifically a point-wise convolution and a depth-wise convolution of stride two and kernel size 5$\times$5.
The convolution block with large stride and kernel size incurs minimal computational overhead while extracts necessary features for the generation of the importance weights.

The importance weights are introduced along the lines of convolutional attention mechanism~\cite{newell16stacked,hu2018squeeze,woo2018cbam}, where attention weights are first determined based on the input instance and then used to recalibrate the activations of certain input dimensions, e.g., channels~\cite{hu2018squeeze}.
In \netName[s], the importance weights are applied to transformation branches of each connection, where each candidate branch is coupled with a respective weight for branch selection.

The Gumbel-Softmax~\cite{jang2016categorical,maddison2016concrete} and reparameterization technique~\cite{kingma14auto} are integrated to recalibrate the importance weights generated by \supernet[s].
The weight recalibration is a continuous relaxation of the \textit{categorical} branch sampling process, which enables tractable gradient-based optimization for the entire network during training.
Specifically, $\widetilde{w}_{c,b}^l$ (the importance weight of $b_{th}$ branch of the $c_{th}$ connection in the $l_{th}$ cell) after the recalibration of ${w}_{c,b}^l$ ($\boldsymbol{W}_c^l \in \mathbb{R}^{B+1}$) with Gumbel-Softmax follows a \textit{Concrete Distribution}~\cite{maddison2016concrete}:

\begin{equation}
    \begin{aligned}
        \widetilde{w}_{c,b}^l = \frac{\exp((w_{c,b}^l + g_{c,b}^l)/\tau)}{  \sum_{b'=0}^{B} \exp((w_{c,b'}^l + g_{c,b'}^l)/\tau)}, \tau>0
    \end{aligned}
    \label{eq:gumbel_softmax}
\end{equation}

\noindent
where $\tau$ is the temperature of the softmax and is annealed steadily during training, and $g_{c,b}^l$=$-\log(-\log(u_{c,b}^l))$ is a $Gumbel(0, 1)$~\cite{maddison2016concrete} random variable for the $b_{th}$ branch by sampling $u_{c,b}^l$ from \textit{Uniform Distribution} $\mathcal{U}(0,1)$~\cite{jang2016categorical}.
$\widetilde{w}_{c,b}^l$ is then directly used for branch weighting in Equation~\ref{eq:connection_op}.
Denoting parameters of the backbone network and \supernet[s] as $\theta$ and $\phi$, then the objective function $\mathcal{L}_{CE}(\theta, \phi)$ can be reparameterized as:

\begin{equation}
    \begin{aligned}
    \mathcal{L}_{CE}(\theta, \phi) &= \mathbb{E}_{\widetilde{w} \sim p_{\phi}(\mathbf{x})} [f_{\theta}(\mathbf{x}, \widetilde{w})] \\ 
    &= \mathbb{E}_{g \sim Gumbel(0,1)} [f_{\theta}(\mathbf{x}, h_{\phi}(\mathbf{x}, g))]
    \end{aligned}
    \label{eq:reparameterization}
\end{equation}

\noindent
where $\mathbf{x}$ is the current input instance and the dependence of the importance weights $\widetilde{w}$ on the parameters of $\phi$ can be transferred from the sampling of $p_{\phi}(\mathbf{x})$ into $h_{\phi}(\mathbf{x}, g)$.
With such reparameterization, the branch selection weights $\widetilde{w}_{c,b}^l$ can be computed as a deterministic function of the weights $w_{c,b}^l$ generated from \supernet[s] with parameters $\phi$ and an independent random variable $u_{c,b}^l$, such that non-differentiable categorical branch sampling process is made directly differentiable with respect to $\phi$ during training.
The weight $\widetilde{w}_{c,b}^l$ following \textit{Concrete Distribution}~\cite{maddison2016concrete} satisfies the nice properties: (1) $\widetilde{w}_{c,b}=\frac{1}{B+1}, \tau \rightarrow +\infty$, and (2)

\begin{equation}
    \begin{aligned}
    p(\lim_{\tau \rightarrow 0} \widetilde{w}_{c,b}^l =1)=\exp(w_{c,b}^l) / \sum_{b'=0}^{B} \exp(w_{c,b'}^l )
    \end{aligned}
    \label{eq:gumbel_p2}
\end{equation}

\noindent
which indicates that the \textit{softmax} computation of Equation~\ref{eq:gumbel_softmax} smoothly approaches discrete \textit{argmax} branch selection as the temperature $\tau$ anneals.
High temperature leads to uniform dense branch selection, while a lower temperature tends to select the most important branch following a \textit{Categorical Distribution} parameterized by $softmax(\boldsymbol{W}_c^l)$.

\subsection{Optimization and Efficient Inference}
\label{sec:sec_optimization}

With the continuous relaxation of the Gumbel-Softmax (Equation~\ref{eq:gumbel_softmax}) and reparameterization (Equation~\ref{eq:reparameterization}), the branch selection process of the \supernet[s] is made directly differentiable with respect to the parameters of \supernet[s] using the chain rule:

\begin{equation}
    \begin{aligned}
    \nabla_{\phi} \mathcal{L}_{CE}(\theta, \phi) &= \mathbb{E}_{g} [\nabla_{\phi} f_{\theta}(\mathbf{x}, h_{\phi}(\mathbf{x}, g))] \\
    &= \mathbb{E}_{g} [f'_{\theta}(\mathbf{x}, h_{\phi}(\mathbf{x}, g)) \nabla_{\phi} h_{\phi}(\mathbf{x}, g)]
    \end{aligned}
    \label{eq:back_propogation}
\end{equation}

\noindent
where the gradient $\frac{\partial \mathcal{L}}{\partial \widetilde{w}_{c,b}^l}$ backpropagated from the loss function $\mathcal{L}_{CE}$ to $\widetilde{w}_{c,b}^l$ through the backbone network via $f'_{\theta}(\mathbf{x}, h_{\phi}(\mathbf{x}, g))$ can be directly backpropagated to $w_{c,b}^l$ with low variance~\cite{maddison2016concrete}, and further to the $\supernet[]^l$ via $\nabla_{\phi} h_{\phi}(\mathbf{x}, g)$ unimpededly.
Therefore, parameters of the entire \netName[] can be optimized in an end-to-end manner by gradient descent effectively.

During training, the temperature $\tau$ of Equation~\ref{eq:gumbel_softmax} regulates the sparsity of the branch selection.
A relatively higher temperature forces the weights to distribute more uniformly so that all the branches can be efficiently trained.
While a low temperature instead tends to sparsely sample one branch from the categorical distribution parameterized by the importance weights dynamically determined by \supernet[s], and thus supports efficient inference by routing inputs to only necessary branches.
We thus propose a two-stage training scheme for \netName[s]:
(1) the first stage pretrains the entire network with a \textit{fixed} relatively high temperature till convergence.
(2) the second stage fine-tunes the parameters with $\tau$ steadily annealing to a low temperature.
The first stage ensures that all the branches are sufficiently trained and specialized for certain input instances; the fine-tuning and annealing in the second stage help maintain the performance of \netName[s] with dynamic routing during inference for higher efficiency.

To further improve inference efficiency, a regularization term is explicitly introduced during the fine-tuning stage, which takes into account the \textit{expectation} of the resource consumption $\mathcal{R}$ in the final loss function $\mathcal{L}$ for the already \textit{correctly-classified} input instances:

\begin{equation}
    \begin{aligned}
        \mathcal{L} &= \mathcal{L}_{CE} +  \lambda \mathds{1}_\mathrm{\hat{\mathbf{y}} = \mathbf{y}} \log \mathbb{E}[\mathcal{R}] \\ 
        &\approx \mathcal{L}_{CE} + \lambda \mathds{1}_\mathrm{\hat{\mathbf{y}} = \mathbf{y}} \log \sum_{l=1}^L \sum_{c=1}^C \sum_{b=0}^B  \widetilde{w}_{c,b}^l \cdot \mathcal{R}(\mathcal{F}_{c,b}^l (\cdot))
    \end{aligned}
    \label{eq:objective_function}
\end{equation}

\noindent
where $\mathcal{L}_{CE}$ denotes the cross-entropy loss (Equation~\ref{eq:reparameterization}), $\mathbf{y}$ is the ground truth class label, $\hat{\mathbf{y}}$ is the prediction, $\lambda$ controls the regularization strength and $\mathcal{R}(\cdot)$ calculates the resource consumption of each branch $\mathcal{F}_{c,b}^l (\cdot)$.
The branch importance weight $\widetilde{w}_{c,b}^l$ represents the probability of the corresponding transformation $\mathcal{F}_{c,b}^l (\cdot)$ being selected during inference, and thus the regularization term $\mathbb{E}[\mathcal{R}]$ corresponds to the expectation of the computational resource required for each input instance.
The resource regularizer can be adapted based on deployment constraints, which may include the parameter size, FLOPs, and memory access cost (MAC), and etc.
In this paper, we mainly focus on the inference time measured by FLOPs, where $\mathcal{R}(\mathcal{F}_{c,b}^l (\cdot))$ is a constant and can be calculated beforehand.
This indicates that the regularizer $\mathcal{R}$ is also directly differentiable with respect to $\widetilde{w}_{c,b}^l$ during the optimization.
We denote \netName[s] trained with regularization strength $\lambda$ as \textit{\netName-R-}$\lambda$.

As illustrated in Figure~\ref{fig:inference}, during inference, dynamic routing for efficiency is achieved by passing the input to the top-$k$ most important branches out of the $B$+1 branches for each connection, whose overall importance weights denoted as $s_{c}^l$ just exceeds a predetermined threshold $T$.
After the selection, the recalibration weight $\widetilde{w}_{c,b}^l$ of the selected branch is rescaled by $\frac{1}{s_{c}^l}$ to stabilize the scale of the connection output.
Consequently, each \supernet[] selects only necessary branches for each instance depending on the input difficulty and the computational cost of each branch to trade off between $\mathcal{L}_{CE}$ and $\mathcal{R}$.
In this paper, the same threshold is shared among all connections for simplicity, and \netName[s] inference with a threshold $t$ is denoted as \textit{\netName-T-t}.

Under such an inference scheme, the backbone network comprises up to $(2^{B+1}-1)^{L\cdot C}$ possible candidate subnets, corresponding to each unique branch selection of all $L\cdot C$ connections. 
For a small 10 cells \textit{\netName}, with 8 connections per cell and 5 candidate operations per connection, there are ${(2^5-1)^{8 \cdot 10}}\approx2\cdot 10^{119}$ possible candidate architectures of different branch selections, which is orders of magnitudes larger than the search space of conventional NAS~\cite{pham2018efficient,liu2018darts,xie2018snas,cai2018proxylessnas,stamoulis2019single}.
\section{Experiments}
\label{sec:experiment}

We mainly focus on the evaluation of the accuracy and efficiency of \netName[s] on benchmark datasets.
We compare the results of \netName[s] against the best-performing expert-designed, NAS-searched and dynamic inference architectures.
Experimental details are provided in Section~\ref{subsec:exps/setup}, and main results are reported in Section~\ref{subsec:evaluation}.
We discuss and visualize the dynamic routing process in Section~\ref{subsec:visualization}.

\subsection{Experimental Setup}\label{subsec:exps/setup}

\powerpoint{Dataset.}
Following conventions~\cite{xie2018snas,liu2018darts}, we report the performance of \netName[s] on benchmark datasets CIFAR-10 and ImageNet-12, where the accuracy and inference efficiency are compared with other state-of-the-art architectures.
CIFAR-10 contains 50,000 training images and 10,000 test images of $32 \times 32$ pixels in 10 classes.
We adopt standard data pre-processing and augmentation pipeline~\cite{liu2018darts,xie2018snas} and apply AutoAugment~\cite{autoaugment}, cutout~\cite{cutout} of length 16.
ImageNet contains 1.2 million training and 50,000 validation images in 1000 classes.
We adopt a standard augmentation scheme following~\cite{liu2018darts,xie2018snas} and apply label smoothing of 0.1 and AutoAugment. Results are reported with a $224\times224$ center crop.

\powerpoint{Temperature Annealing Scheme.}
In the pre-training stage, the temperature $\tau$ is fixed to 3 till convergence.
In the fine-tuning stage, $\tau$ is reset to 1.0 and is further annealed by $\exp(-0.0006) \approx 0.999$ every epoch to 0.5.
The initial $\tau$ is 5 and exponentially annealed to 0.8 for ImageNet.

\powerpoint{Candidate Operation Set.}
The following 5 candidate operations ($B$+1=5) are adopted for demonstration, which can be readily adjusted in deployment:

{\setlength\multicolsep{0pt}
\begin{multicols}{2}
\begin{itemize}[noitemsep,topsep=0pt]
  \item $3 \times 3$ max-pooling
  \item $3 \times 3$ avg-pooling
  \item skip connection
  \item $3 \times 3$ separable-conv
  \item $5 \times 5$ separable-conv
  \item[\vspace{\fill}]
\end{itemize}
\end{multicols}}

In particular, skip connection is adopted as the fast-forward branch that allows for efficient input forwarding; pooling layers contain no parameter and are computationally lightweight; separable-conv dominates the parameter size and computation in each connection, which contains separable convolutions of \textit{ReLU-Conv-Conv-BN}.
The three types of transformations support \textit{trade-offs} between model capacity and efficiency for the branch selection of each connection.

\powerpoint{Architecture Details}\label{sec:archi_details}
We adopt three \netName[] architectures of different size:
(1) \netName(S), a smaller network with $L$=5 cells and 15 initial channels;
(2) \netName(M), a medium network with 10 cells and 20 initial channels;
(3) \netName(L), a larger network with 10 cells and 32 initial channels.

All the architectures adopt $N$=$4$ intermediate nodes for each cell and a plain node connection strategy where each node is connected to $n$=$2$ preceding nodes, specifically $x_{i-1}^l$ and randomly $x_{0}^l$ or $x_{1}^l$ for simplicity.
Further, nodes directly connected to input nodes are downsampled with stride 2 for the $\frac{L}{3}$-th and $\frac{2L}{3}$-th cells.
An auxiliary classifier with weight 0.4 is connected to the output of the $\frac{2L}{3}$-th cell for additional regularization.

\powerpoint{Optimization Details.}
Following conventions~\cite{liu2018darts,xie2018snas,shu2019understanding}, we apply SGD with Nesterov-momentum 0.9 and weight decay $3 \cdot 10^{-5}$ for 250 epochs of 0.97 learning rate decay for ImageNet.
For CIFAR-10, we apply SGD with momentum 0.9 and weight decay $3\cdot10^{-4}$ for 1200 epochs for both training stages. 
The learning rate is initialized to 0.025 and 0.005 for the pre-training and fine-tuning stage respectively.
The batch size is set to 128 to fit the entire \netName[] into one NVIDIA Titan RTX.
We adopt higher structural level dropout for better regularization during training, specifically, drop-connection linearly increased to 0.1, and drop-branch to 0.7 for CIFAR-10.
The learning rate is annealed to zero with the cosine learning rate scheduler~\cite{ilya17sgdr}.

\subsection{Performance Evaluation}\label{subsec:exps/evaluation}
\label{subsec:evaluation}

\paragraph{Overall Results and Discussion.}

\renewcommand{\arraystretch}{1.1}
\begin{table*}[ht]
\centering
\resizebox{\textwidth}{!}{
\begin{tabu}{ L{3.8cm}||C{2.2cm}C{2.2cm}C{2.2cm}|C{2.5cm}C{2.5cm}C{2.4cm}}
 \tabucline[1.5pt]{-}
    \multirow{2}{*}{\textbf{Architecture}} & \textbf{Test Error} & \textbf{Params} & \textbf{FLOPs} & \multirow{2}{*}{\textbf{Search Method}} & \multirow{2}{*}{\textbf{Search Space}} &  \textbf{Search Cost} \\ 
     
    & \textbf{(\%)} & \textbf{(M)} & \textbf{(M)} &  &  & \textbf{(GPU days)} \\
    \tabucline[.5pt]{-}
    
    ResNet-110~\cite{he2016deep} & 6.43 & 1.73 & 255.3 & manual & -- & --\\
    DenseNet-L190-k40~\cite{huang2017densely} & 3.46 & 25.6 & 9345 & manual & -- & -- \\
    ShuffleNetV2 1.5$\times$~\cite{ma2018shufflenet}$\dagger$ & 6.36 & 2.49 & 95.70 & manual & -- & -- \\
    MobileNetV2 1.0$\times$~\cite{sandler2018mobilenetv2}$\dagger$ & 5.56 & 2.30 & 94.42 & manual & -- & -- \\
    
    \tabucline[.5pt]{-}
    NASNet-A~\cite{zoph2018learning}$\dagger$ & 2.65 & 3.3 & 505.1 & RL & cell & 1800 \\
    AmoebaNet-A~\cite{real2019regularized}$\dagger$ & 3.12 & 3.1 & 514.9 & evolution & cell & 3150 \\
    ENAS~\cite{pham2018efficient}$\dagger$ & 2.83 & 4.7 & 767.8 & RL & cell & 2.2 \\
    DARTS~\cite{liu2018darts} & 3.00 & 3.3 & 542.0 & gradient & cell & 1.9 \\
    \tabucline[.5pt]{-}
    ConvNet-AIG-110~\cite{aig} & 5.76 & 1.78 & 410 & gradient & layer-wise & --\\
    SkipNet-110-HRL~\cite{wang2018skipnet} & 6.70 & 1.75 & 150.6 & RL & layer-wise & -- \\
    MSDNet~\cite{huang2017multi}$\dagger$ & 6.82 & 5.44 & 54.35 & manual & -- & -- \\
    
    \textbf{\netName(S)-R-0.1-T-0.8} & 3.66 / 4.21 & \textbf{0.57 / 0.31} & \textbf{84.65 / 43.82} & gradient & layer-wise & -- \\
    \textbf{\netName(M)-R-0.1-T-0.8} & 2.84 / 3.44 & 1.86 / 0.89 & 267.3 / 119.6 & gradient & layer-wise & -- \\
    \textbf{\netName(L)-R-0.1-T-0.8} & \textbf{2.27 / 2.84} & 4.41 / 2.16 & 603.1 / 265.8 & gradient & layer-wise & -- \\
     
\tabucline[1.5pt]{-}
\end{tabu}}
\caption{
Performance (Inference with All Branches/Dynamic Routing) of \netName[s] compared with representative expert-designed, NAS-searched and dynamic inference architectures on CIFAR-10.
Results marked with $\dagger$ are obtained by training respective architectures with our implementation.
}
\label{tab:comparison}
\vspace{-3mm}
\end{table*}

Table~\ref{tab:comparison} summarizes the overall performance of \netName[s] on CIFAR-10.
In terms of total training cost, \netName[s] take only 2.5 and 5.5 GPU training days for \netName(S) and \netName(M) respectively without explicit architecture search.
The training time of \netName[] is up to three orders of magnitudes less than evolution-based NAS or RL-based NAS, thanks to the efficient end-to-end gradient-based optimization scheme.

\renewcommand{\arraystretch}{1.}
\begin{table}[h!]
\centering
\resizebox{0.45\textwidth}{!}{
\begin{tabu}{ L{4cm}||C{1.8cm}C{1.5cm}C{1.5cm}}
    \tabucline[1.5pt]{-}
    
    \multirow{ 2}{*}{\textbf{Architecture}} & \textbf{Test Error} & \textbf{Params}& \textbf{FLOPs}\\
    & \textbf{(\%)} & \textbf{(M)} & \textbf{(M)}\\

    \hline
    1.0 MobileNet~\cite{howard2017mobilenets} & 29.4 & 4.2 & 569 \\
    ShuffleNet $2\times$~\cite{zhang2018shufflenet} & 29.1 & 5.0 & 524\\
    \hline
    
    NASNet0-A~\cite{zoph2018learning} & 26.0 & 5.3 & 564   \\
    DARTS~\cite{liu2018darts} & 26.9 & 4.9 & 595    \\
    
    \hline
    
    ConvNet-AIG-50-t-0.4~\cite{aig}	& 24.8	& 26.56	& 2560  \\
    SkipNet-101~\cite{wang2018skipnet} &	27.9 &	26.56	&	2147    \\

    \netName(L)-R-0.01-T-0.8 & 29.8 & 3.8  & 351  \\
    
    \tabucline[1.5pt]{-}
\end{tabu}}
\caption{
Inference performance of \netName(L) compared with representative expert-designed, NAS-searched and dynamic inference architectures on ImageNet.
}
\label{tab:imagenet}
\vspace{-3mm}
\end{table}

For inference performance, \netName[s] with dynamic routing considerably reduce the parameter size and FLOPs compared with baseline networks.
In particular, \netName(S)-R-0.1-T-0.8 takes 0.31M parameters and 43.82M FLOPs on average during inference, specifically, only 13.48\% and 46.41\% of efficient network MobileNetV2 $1.0\times$~\cite{sandler2018mobilenetv2} with 1.35\% higher accuracy; \netName(M)-R-0.1-T-0.8 achieves up to 3.71x and 4.53x parameter size and FLOPs reduction compared with DARTS~\cite{liu2018darts}, with a minor 0.44\% accuracy decrease; further, \netName(L)-R-0.1-T-0.9 achieves accuracy comparable to the best NAS-searched architectures (2.27\% with all branches and 2.84\% with dynamic routing) while takes only 265.8M Flops, which is around half of their inference FLOPs.
Compared with other dynamic inference networks, in particular, SkipNet-110-HRL~\cite{wang2018skipnet} that dynamically determines to skip each residual layer with an RL-trained hypernetwork,
and MSDNet~\cite{huang2017multi} that supports inference time accuracy-efficiency trade-offs by giving predictions with intermediate features, \netName(S)-R-0.1-T-0.8 achieves a notably lower test error rate of 4.21\% and meanwhile much more efficient inference with only 43.82M FLOPs on average.

We note that such a significant reduction in prediction parameter size and FLOPs can be ascribed to the adoption of efficient operations following MobileNets and NAS, and the fact that not all instances need an expensive architecture to be correctly classified.
Therefore, dynamic routing inputs to only necessary branches can reduce redundant computation considerably.

\renewcommand{\arraystretch}{1.1}
\begin{table*}[h!]
\centering
\resizebox{0.68\textwidth}{!}{
\begin{tabu}{ L{4.5cm}||C{2.4cm}C{3.2cm}C{3.2cm}}
    \tabucline[1.5pt]{-}
    
    \multirow{ 2}{*}{\textbf{Architecture}} & \textbf{Test Error} & \textbf{Params}& \textbf{FLOPs}\\
    & \textbf{(\%)} & \textbf{(M)} & \textbf{(M)}\\

    \hline
    \netName(S)-w/o-\supernet & 4.78 & 0.46 & 77.34                \\
    \hline
    
    \netName(S)-Softmax & 4.37 & 0.57 & 84.65 \\
    \quad $\hookrightarrow$ \netName(S)-Softmax-T-0.8 & 5.27 (+0.90\%) & 0.55 (-3.51\%) & 80.99 (-4.32\%)  \\
    
    \hline
    
    \netName(S)-Gumbel-Softmax & 3.66 & 0.57 & 84.65  \\
    \quad $\hookrightarrow$ \netName(S)-R-0.0-T-0.8 & \textbf{4.07} (+0.41\%) & 0.33 (-42.1\%) & 47.91 (-43.4\%)  \\
    \quad $\hookrightarrow$ \netName(S)-R-0.1-T-0.8 & 4.21 (+0.55\%) & 0.31 (-45.6\%) & 43.82 (-48.2\%) \\
    \quad $\hookrightarrow$ \netName(S)-R-0.5-T-0.8 & 5.85 (+2.19\%) & \textbf{0.20 (-64.9\%)} & \textbf{29.28 (-65.4\%)} \\
    
    \tabucline[1.5pt]{-}
\end{tabu}}
\caption{
Inference performance of \netName(S) with different regularization strengths.
The amount of reduction compared with the respective full networks is given in parentheses.
}
\label{tab:ablation}
\vspace{-3mm}
\end{table*}

Table~\ref{tab:imagenet} reports the performance of representative architectures and \netName[s] with a thresholds of 0.8 on ImageNet.
The results show that \netName(L) obtains competitive prediction performance compared with expert-designed architectures and lower accuracy than NAS-searched architectures.
The lower accuracy can be largely attributed to the simple connection topology adopted in the backbone of \netName, as we focus on showing the effectiveness of the proposed dynamic routing mechanism in reducing unnecessary computation instead of obtaining state-of-the-art accuracy.

With the dynamic routing of \supernet[], one single model of \netName[] supports accuracy-efficiency trade-offs by simply controlling the importance threshold $T$ during inference.
In particular, \netName(L) inference with a threshold 0.8 reduces FLOPs by 27.63\% with a minor accuracy decrease, and can further reduce FLOPs with a lower threshold.
These results show that \netName[s] can also support applications that require runtime accuracy-efficiency control.

\noindent
\textbf{Ablation studies.}
We further examine the effect of the hypernetworks \supernet[s] and \textit{regularization strength} quantitatively in Table~\ref{tab:ablation}.
We train \netName[s] without hypernetworks (\netName(S)-w/o-\supernet), which increases test error by 1.12\% from 3.66\% to 4.78\% as compared with \netName[s] trained with Gumbel-Softmax (\netName(S)-Gumbel-Softmax).
We also find that \netName[s] trained with Gumbel-Softmax obtain noticeably lower test error compared with \netName[s] train with Softmax only (\netName(S)-Softmax).
Further, \netName(S)-Softmax only reduces a very limited amount of 4.32\% FLOPs with dynamic routing inference.
These findings suggest that \supernet[] and the Gumbel-Softmax recalibration are essential to obtain high accuracy and efficiency during inference.
Specifically, \supernet[s] enable instance-aware branch selection, and Gumbel-Softmax makes the selection process optimizable during training and efficient during inference.
We also train \netName(S) with different regularization strengths.
Results show that a larger regularization strength effectively trades off prediction accuracy for higher efficiency.
E.g., with a regularization strength 0.5, \netName[s]-R-0.5-T-0.8 only takes 29.28M inference FLOPs, a reduction of 65.4\% FLOPs, while the test error increases by 1.78\% compared with the full network \netName(S)-Gumbel-Softmax.

\subsection{Visualization of Dynamic Routing}
\label{subsec:visualization}

\begin{figure}[t!]
    \centering
    \begin{subfigure}[t]{0.4\textwidth}
        \centering
        \includegraphics[width=1\textwidth]{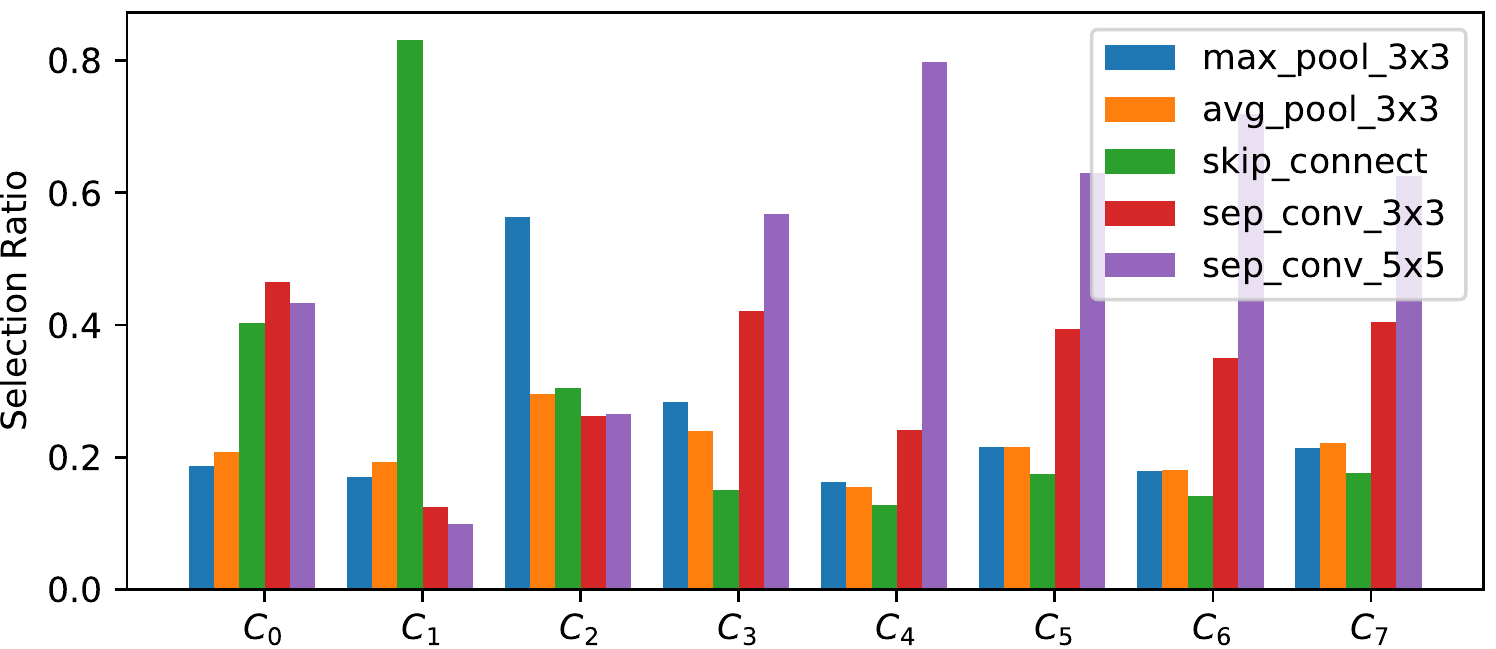}
        \label{fig:weight_middle}
    \end{subfigure}%
    \caption{\textit{Branch selection ratio} for each connection with dynamic routing during inference in a representative cell of \netName(S)-R-0.1-T-0.8 on CIFAR-10.}
    \label{fig:selection_ratio}
    \vspace{-5mm}
\end{figure}

\noindent
\textbf{Branch Selection Ratio.}
We visualize in Figure~\ref{fig:selection_ratio} the average \textit{branch selection ratio} of one representative cell of \netName(S)-R-0.1-T-0.8 with dynamic routing, which shows the ratio of each branch being selected by \supernet[s] during inference.
Figure~\ref{fig:selection_ratio} confirms that the transformations required for different input instances vary greatly.
In particular, lightweight pooling and skip connection branches are commonly selected, and thus, a considerable amount of computation can be saved.


\begin{figure}[t!]
    \centering
    \begin{subfigure}[r]{0.23\textwidth}
        \centering
        \includegraphics[width=0.8\textwidth]{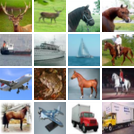}
        \caption{Easy Instances}
        \label{fig:easy}
    \end{subfigure}%
    ~
    \begin{subfigure}[l]{0.23\textwidth}
        \centering
        \includegraphics[width=0.8\textwidth]{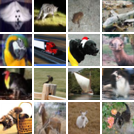}
        \caption{Hard Instances}
        \label{fig:hard}
    \end{subfigure}
    \caption{Visualization of \textit{easy} and \textit{hard} instances of model \netName(M)-R-0.1-T-0.8 on CIFAR-10.
    Easy instances are generally clearer and brighter, while hard instances are darker and blurry.}
    \label{fig:images}
    \vspace{-5mm}
\end{figure}

\noindent
\textbf{Qualitative Difference between Instances.}
Denoting instances that the network is confident with in prediction as \textit{easy instance} and uncertain about as \textit{hard instance}, we can then show the cluster of \textit{easy} and \textit{hard} instances in Figure~\ref{fig:images} to help understand the dynamic routing mechanism.
We find that in \netName[s], the confidence of the prediction is highly correlated to the image quality.
Specifically, easy instances are more salient (clear with high contrast) while hard instances are more inconspicuous (dark with low contrast).
We then compute the accuracy and average FLOPs of these two types of instances.
Easy instances achieve higher classification accuracy with 23.1\% fewer FLOPs on average compared to hard instances.
This suggests that although instances from the same dataset are generally regarded as i.i.d., the prediction difficulty of different instances still differs greatly, and thus a sizeable amount of computation can be reduced by dynamically cutting off unnecessary branches for relatively easier instances.

\section{Conclusion}
\label{sec:conclusion}

In this paper, we present \netName[s], a general architecture framework that supports input-aware inference by dynamic routing.
Lightweight hypernetwork \supernet[s] are integrated to dynamically produce customized architectures for different instances so that inputs can be dynamically routed to only necessary branches for efficiency.
We also introduce Gumbel-Softmax and the reparameterization trick to the routing process, which enables tractable and effective gradient-based training, and more importantly, extremely efficient inference.
The inference efficiency is enhanced with the resource-aware regularization.
Experimental results with ablation studies and visualizations confirm the efficiency of the dynamic routing architecture.

\newpage

{\small
\bibliographystyle{ieee_fullname}
\bibliography{egbib}
}

\end{document}